\documentclass[a4paper,conference]{IEEEtran}
\ifCLASSINFOpdf
\else

\fi

\usepackage{amsmath}
\usepackage{graphicx}
\usepackage[usestackEOL]{stackengine}
\usepackage{caption}
\usepackage{bm}
\usepackage[linesnumbered, ruled]{algorithm2e}
\SetKwRepeat{Do}{do}{while}%
\usepackage{caption}
\usepackage{subcaption}
\usepackage{cite}
\usepackage{float}
\usepackage{tikz}

\usepackage{booktabs}
\usepackage{textcomp}
\SetKwInput{KwData}{Input}
\SetKwInput{KwResult}{Output}
\SetKwRepeat{Do}{do}{while}%

\begin{document}
\title{Robust Adaptive Median Binary Pattern for noisy texture classification and retrieval}

\author{\IEEEauthorblockN{Mohammad Alkhatib and Adel Hafiane}
\IEEEauthorblockA{INSA Centre Val de Loire, Universit\'e d'Orl\'eans, Laboratoire PRISME EA 4229, 
Bourges F-18000, France}

}

\maketitle

\begin{abstract}
Texture is an important cue for different computer vision tasks and applications. Local Binary Pattern (LBP) is considered one of the best yet efficient texture descriptors. However, LBP has some notable limitations, mostly the sensitivity to noise. In this paper, we address these criteria by introducing a novel texture descriptor, Robust Adaptive Median Binary Pattern (RAMBP). RAMBP based on classification process of noisy pixels, adaptive analysis window, scale analysis and image regions median comparison. The proposed method handles images with high noisy textures, and increases the discriminative properties by capturing microstructure and macrostructure texture information. The proposed method has been evaluated on popular texture datasets for classification and retrieval tasks, and under different high noise conditions. Without any train or prior knowledge of noise type, RAMBP achieved the best classification compared to state-of-the-art techniques. It scored more than $90\%$ under $50\%$ impulse noise densities, more than $95\%$ under Gaussian noised textures with standard deviation $\sigma = 5$, and more than $99\%$ under Gaussian blurred textures with standard deviation $\sigma = 1.25$. The proposed method yielded competitive results and high performance as one of the best descriptors in noise-free texture classification. Furthermore, RAMBP showed also high performance for the problem of noisy texture retrieval providing high scores of recall and precision measures for textures with high levels of noise.

\end{abstract}

\IEEEpeerreviewmaketitle

\section{Introduction}
Texture is a fundamental characteristic of various types of images. Texture analysis is considered as a complex problem due to the large number of texture classes, the associated intraclass variations such as randomness and periodicity, and external class variation such as illumination and noise~\cite{tuceryan1993texture}. Texture classification is one of the major problems in texture analysis, it has a crucial value in the fields of computer vision and pattern recognition, including medical imaging, document analysis, environment modeling, and object recognition~\cite{petrou2006image}.

One of texture classification methods that gained huge attention is Local Binary Pattern (LBP). LBP is a powerful operator that shows robustness to illumination, rotation and scale~\cite{ojala2002multiresolution}. And due to its low computational complexity, LBP has been used widely as a solution for many problems, such as texture classification~\cite{liao2009dominant}, object detection ~\cite{satpathy2014lbp}, image matching~\cite{heikkila2009description},  image retrieval~\cite{doshi2012comprehensive}, biomedical image analysis~\cite{nanni2010local}, face recognition~\cite{ahonen2004face}, etc. For general texture classification purposes, LBP derivatives have been introduced such as ILBP~\cite{jin2004face}, CLBP~\cite{guo2010completed}, RLBP~\cite{chen2013rlbp}, DLBP~\cite{liao2009dominant}, etc.

However, LBP and its derivatives have their weaknesses in term of robustness to noise. For that, several studies aimed to present a noise robust operator. In~\cite{hafiane2007median}, the authors proposed Median Binary Pattern (MBP) to add more sensitivity to microstructure and impulse noise robustness. Nevertheless, MBP does not handle other types of noise and showed low performance for the high levels of impulse noise. In~\cite{liu2014brint}, the authors introduced Binary Rotation Invariant and Noise Tolerant (BRINT). Although BRINT samples the points in a scaled circular neighborhood which made it more distinctive and robust to noise, it suffers from limitations in term of robustness under high noisy textures. 

In~\cite{schaefer2012multi}, Schaefer et al. proposed Multi-Dimensional Local Binary Pattern (MDLBP), which added more information from different radii and concatenated it in one histogram. This makes the histogram more effective. But this approach suffers in robustness under high noise corruption and from computational complicity due to feature dimensionality. In~\cite{hafiane2015joint}, the authors introduced Adaptive Median Binary Pattern (AMBP). AMBP used self-adaptive analysis window size depending on the local microstructure of the texture which made the descriptor more robust to impulse noise. Despite the noise robustness of this approach, it has limitations for textures with a very high level of noise. 

In~\cite{cimpoi2015deep}, Cimpoi et al. used filter banks and convolutional neural networks (CNN) for texture recognition and segmentation. The descriptor built on pre-trained VGG-VD (very deep) model which improves the performance. This approach is an effective texture descriptor and produces more robustness for variable images recognitions, but it has weaknesses to the median and high noisy textures, and some shortcomings in term of time complexity.   

\begin{figure*}[t]
\centering
\includegraphics[width=6.5in]{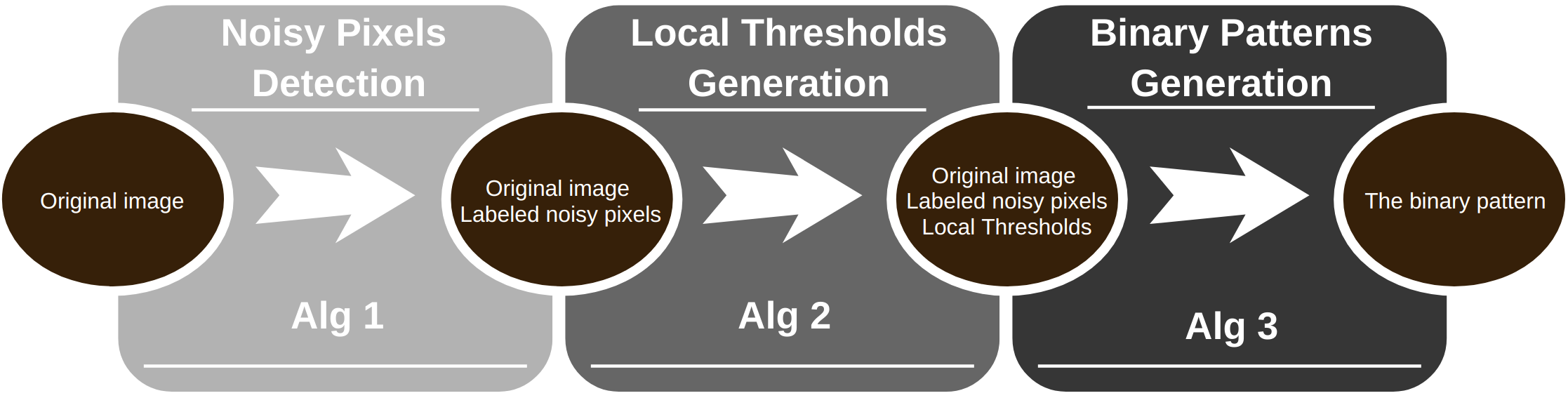}
\DeclareGraphicsExtensions.
\caption{Illustration for the proposed descriptor.}
\label{fig 1}
\end{figure*}

Guo et al.~\cite{guo2016robust} proposed Scale Selective Local Binary Patterns (SSLBP). SSLBP uses a Gaussian filter to produce a scale space of a texture image. For each image in the scale space, the pre-learned dominant binary patterns histogram is built. Then the scale invariant feature for each pattern is found by taking the maximal frequency among different scales. SSLBP considered an effective descriptor for textures under Gaussian noise. Nevertheless, SSLBP filtering procedure failed under impulse noise. 

Liu et al.~\cite{liu2016median} performed median filtering and compared it with a sampling scheme to introduce Median Robust Extended Local Binary Pattern (MRELBP). MRELBP adds more microstructure and macrostructure information, but it used simple median filter procedure which leads to failure under Gaussian noise and extremely high level of impulse noise. 

Although binary patterns family has huge success in the computer vision field, there are several weaknesses with these methods. In~\cite{liu2017local}, the authors performed extensive comparisons for the existing local binary features for texture classification, where many of the existing local binary approaches suffer from a serious limitation. These limitations can be concluded in the descriptor ability to handle textures with a high level of noise, and to handle different types of noise. 

Based on previous descriptors limitations, the obvious question being raised here is how to reach high noise robustness without any prior knowledge of the noise type, without any prior learning process, and for different kind of noises such as Gaussian noise, Gaussian blur, and impulse noise. In other words, performing the descriptor in noise-free data then try to classify and retrieve the noisy and noise-free textures under different geometric and illumination condition. 
 
Gaussian noise, Gaussian blur and impulse noise are considered as the most frequent and challenging noises in image processing, computer vision and pattern recognition fields. For that, to improve and ensure the best performance of the image processes such as classification, these noises should be detected, reduced or removed. Some descriptors incorporate filtering procedure to improve the performance, such as Gaussian filtering for SSLBP, and median filtering for MBP, AMBP and MRELBP. Many techniques have been developed to suppress Gaussian noise, such as mean filter, wavelet denoising~\cite{figueiredo2007majorization}, and kernel regression~\cite{takeda2007kernel}. Nevertheless, these filters are suitable for Gaussian noise but not for other noises such as impulse noise. On the other hand, various filters have been proposed to remove impulse noise, such as median filter~\cite{buades2005review} and adapted median filter~\cite{hwang1995adaptive}.  Hence median and adaptive median filters consider all pixels as noisy corrupted pixels, the filter will fail under images with a high level of noise. To avoid this drawback, the switching techniques were introduced such as Boundary Discriminative Noise Detection (BDND), which takes the advantages of detecting which pixel is corrupted and which one is not~\cite{ng2006switching,eng2001noise,zhang2009impulse}. In this context, using pixel classification from switching techniques with binary pattern methods can lead to better texture analysis for different types of noise.

In this paper, we propose an efficient and simple local binary descriptor, Robust Adaptive Median Binary Pattern (RAMBP). It takes the advantages of switching techniques and median adaptive scheme to include more robustness in features for texture with a high level of noise. RAMBP captures both microstructure and macrostructure texture information, and provides a better representation of the local structures. RAMBP effectiveness and robustness will be examined for high noisy textures classification and retrieval. 

The major contributions of this paper can be summarized as follows: 
\begin{itemize}
   \item Robust descriptor for different high noises.
   \item High noisy texture classification.
   \item Noise-free texture classification.
   \item Noisy texture retrieval.
\end{itemize}

The structure of our paper is as follow. Section~\ref{sec:PA} details the proposed RAMBP method. Followed by experimental results and discussion in Section~\ref{sec:EX}. The paper is ended with final conclusions in Section~\ref{sec:Con}.

\section{The proposed approach}
\label{sec:PA}

To provide an efficient texture classification process, the descriptor should be discriminative and robust to noise. All state-of-the-art descriptors share one or more weaknesses of sensitivity for high noisy textures. RAMBP uses noisy pixel classification, an adaptive window for the threshold and binary modules, and regional values instead of using pixels intensities. Fig.~\ref{fig 1} shows the scheme of the proposed descriptor, where it can be seen it's divided into three stages, classification process of noisy pixels detection, threshold process, and generating the binary pattern.

\subsection{Classification process for noisy pixels detection}
\label{sec:PC}

As this paper aims to perform texture classification without any prior noise knowledge, the first step consists of classifying each pixel in the image as corrupted or uncorrupted pixels. for that, the detection step of BDND algorithm~\cite{ng2006switching} has been adopted in this paper.

Pixel classification starts by taking a $21\times21$ window around the central pixel, then examine the pixel whether it meets the condition as an uncorrupted pixel. If the pixel considered as a corrupted pixel in the first stage, another examination will be invoked by imposing a $3\times3$ window around the central pixel to ensure the examination for more confined local statistics. A pixel classified as a corrupted pixel, if it fails in both examinations. Alg.~\ref{Alg1} provides a full explanation about pixel classification step.

\begin{figure}[t]
\centering
\includegraphics[width=2in]{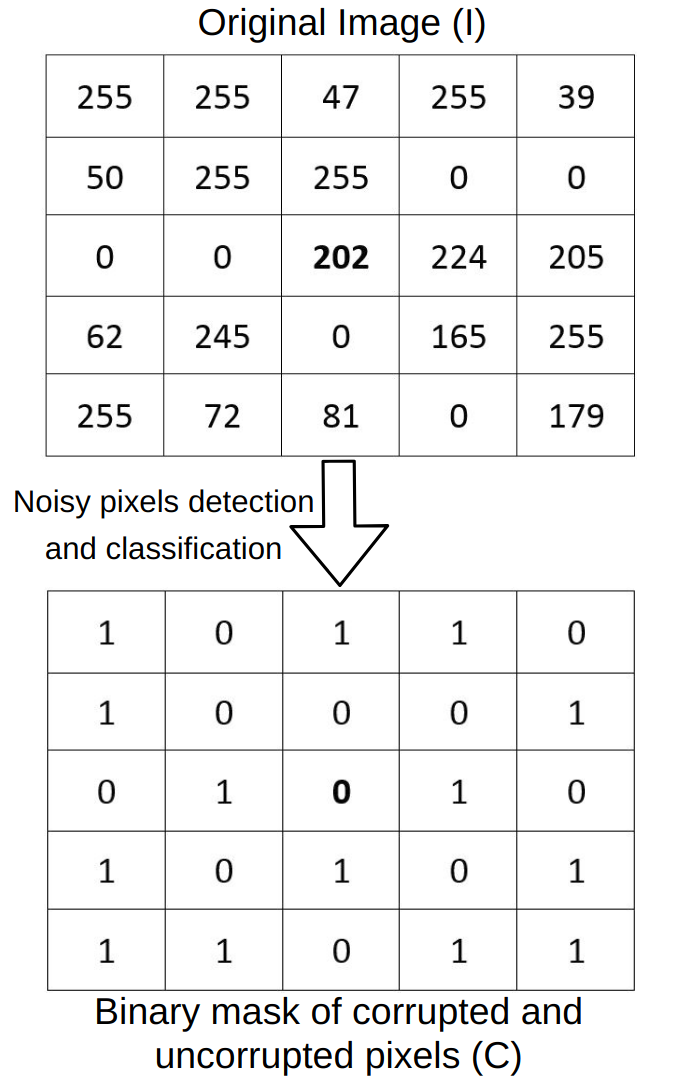}
\DeclareGraphicsExtensions.
\caption{Pixel classification example of $5\times5$ window using the procedure of Alg.~\ref{Alg1}. In the image $C$, corrupted pixels are represented as $0$ and uncorrupted pixels as $1$.}
\label{fig 2}
\end{figure}

Fig.~\ref{fig 2} provides an example of $5\times5$ window instead of $21\times21$ to facilitate understanding pixel classification algorithm (Alg.~\ref{Alg1}) using the following procedure,
\begin{itemize}
\item The first step after choosing the window is to sort all the pixels in the window to obtain $V_0$, in the given example (Fig.~\ref{fig 2}) $V_0 = [0\; 0\; 0\; 0\; 0\; 0\; 39\; 47\; 50\; 62\; 72\; 81\; \boldsymbol{165}\; $ $179\; 202\; 205\; 224\; 245$ $255$ $255\; 255\; 255\; 255\; 255\; 255]$, and the median value ($med$) of $V_0$ is $\boldsymbol{165}$. 
\item Then the difference vector is obtained $D_V = [0$ $0$ $0$ $0$ $0$ $39$ $8$ $3$ $12$ $10$ $9$ $\boldsymbol{84}$ $14$ $\boldsymbol{23}$ $3$ $19$ $21$ $10$ $0$ $0$ $0$ $0$ $0$ $0]$.
\item Find $v_L$, the correspondence pixel in $V_0$ that gives the maximum intensity differences in $D_V$ left interval. (left interval of $D_V$ is located between the $0$ and $med$ in $V_0$). In this example $v_L = \boldsymbol{81}$.
\item In the same manner, find $v_R$, the correspondence pixel in $V_0$ that gives the maximum intensity differences in $D_V$ right interval. (right interval of $D_V$ is located between the $med$ and $255$ in $V_0$). And in this example $v_R = \boldsymbol{179}$.
\item Then the three clusters are \big\{${0,0,0,0,0,0,39,47,50}$ ${,62,72,81}$\big\}, \big\{${165,179}$\big\}, and \big\{${202,205,224,245,255,}$ ${255,255,255,255,255,255,255}$\big\}. The central pixel $I(x_0) = \boldsymbol{202}$ belongs to the third cluster which considered as corrupted pixel and the pixel needs to re-examine on $3\times3$ window .
\item As can be seen in Fig.~\ref{fig 1}, sorting the pixels in the $3\times3$ window gives $V_0 = [0$ $0$ $0$ $165$ $\boldsymbol{202}$ $224$ $245$ $255$ $255]$ with $med = \boldsymbol{202}$. $D_V = [0$ $0$ $\boldsymbol{165}$ $\boldsymbol{37}$ $22$ $10$ $0]$. $v_L = \boldsymbol{0}$ and $v_R = \boldsymbol{165}$, which provide three clusters \big\{${0,0,0}$\big\} , \big\{${165}$\big\}, and \big\{${202,224,245,255,255}$\big\}. The central pixel $I(x_0) = \boldsymbol{202}$ still belongs to the corrupted pixels clusters, which concludes that is a corrupted pixel and $C(x_0) = \boldsymbol{0}$.
\end{itemize}

\begin{algorithm}
  \KwData{The original image $I$.}
  \KwResult{The image of labeled pixels $C$.}
\For{each pixel position \boldsymbol{$x_0$} }{ 
For the current pixel $I(x_0)$, impose a $21 \times 21$ window.\\
Compute $V_0$ by sorting the pixels in the window.\\
Find the median value($med$) of $V_0$.\\ 
From $V_0$, obtain the difference vector $D_V$. $D_V[i] = V_0[i+1] - V_0[i]$, where the index $i = 1:length(V_0)-1$\\
Compute the left cluster range $v_L$, where $v_L = V_0[i_l]$ ($i_l$ is the index of $max(D_V\big\{{0,i_{med}}\big\})$).\\
Compute the right cluster range $v_R$, where $v_R = V_0[i_r]$ ($i_r$ is the index of $max(D_V\big\{{i_{med},end}\big\})$).\\
Initialize three clusters of $V_0$, \big\{${0,v_L}$\big\}, \big\{${v_L,v_R}$\big\}, and \big\{${v_R,255}$\big\}.\\
\uIf{ $I(x_0)$ $\in$ \big\{${v_L,v_R}$\big\}}{
$x_0$ labeled as $uncorrupted$ $pixel$.\\
}
 \Else{
    Repeat $2-8$ steps with $3\times3$ window around $x_0$. If $I(x_0)$ $\notin$ \big\{${v_L,v_R}$\big\}, label $x_0$ as $corrupted$ $pixel$. Otherwise, $x_0$ labeled as $uncorrupted$ $pixel$. \\
  }
}
  \caption{Noisy pixels detection}
  \label{Alg1}
\end{algorithm}

\subsection{Threshold Process}
\label{sec:TP}

\begin{figure}[t]
\centering
\includegraphics[width=2in]{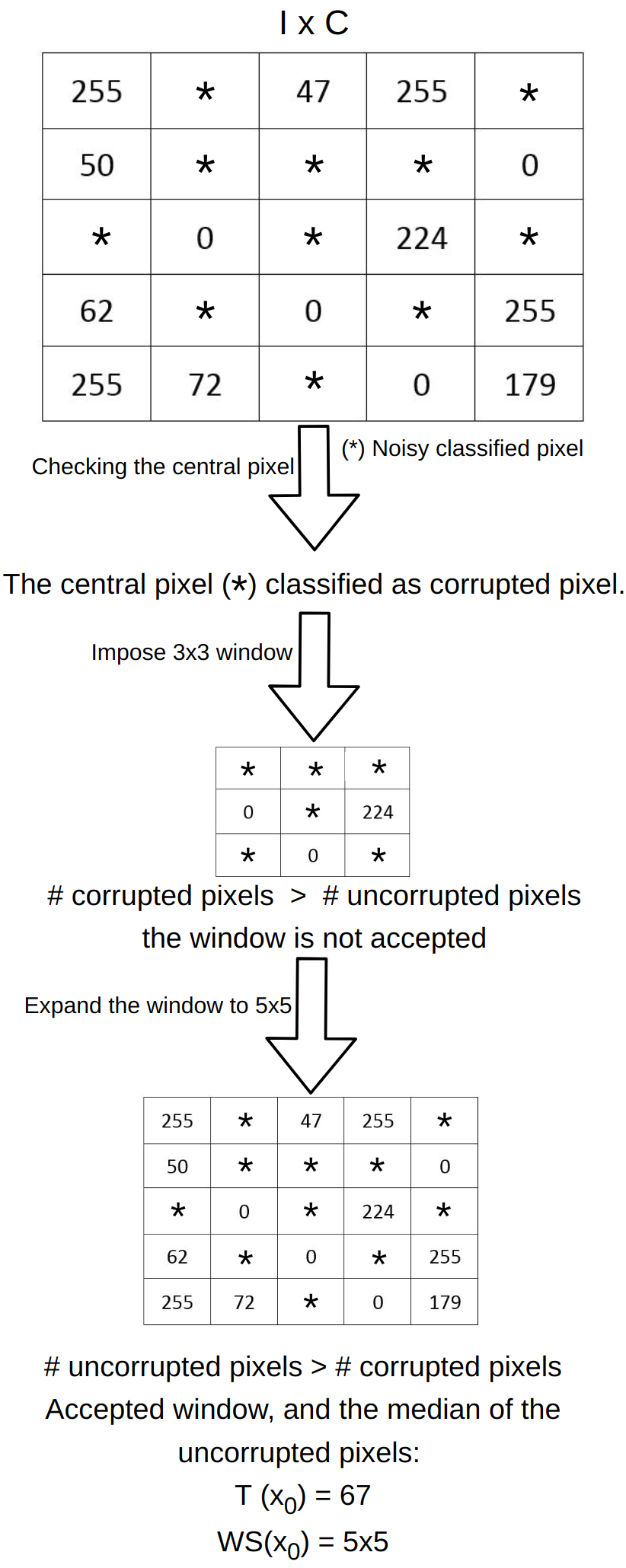}
\DeclareGraphicsExtensions.
\caption{Threshold process example to facilitate understanding Alg.~\ref{Alg2}.}
\label{fig 3}
\end{figure}

Finding the threshold value for each pixel is a crucial point for generating the binary pattern. Using a corrupted central pixel as a threshold value, as LBP, will affect the noise robustness of the descriptor. As well as, using a small or large region to obtain the median as a threshold value will affect the descriptor. This leads to biased median value, due to missing information for the small region or including a large number of pixels for the large region. For obtaining the threshold value, adaptive window and pixel classification are used to reach the maximum robustness.

Alg.~\ref{Alg2} represents the threshold process of the proposed descriptor, which starts by checking if the current pixel is classified as a corrupted or an uncorrupted pixel. If the current pixel classified as an uncorrupted pixel, the pixel threshold value is equal to the current pixel value (same as LBP). Otherwise, a $3\times3$ window is imposed around the current pixel and the number of the uncorrupted pixels is counted. If the number of the uncorrupted pixel is more than the corrupted ones, the threshold value is equal to the median of the uncorrupted pixels inside this window. Otherwise, the window will be enlarged by $1$ pixel in all directions ($5\times5$). This process will be repeated until the maximum window size is reached, where the threshold value is equal to the median value of all uncorrupted pixel inside that window.

Fig.~\ref{fig 3} illustrates an example of obtaining the threshold value after classifying the pixels using the following procedure,
 
\begin{itemize}
\item The current central pixel is classified as a corrupted pixel, which leads to impose a $3\times3$ window around it.
\item The next step consists of checking whether the number of uncorrupted pixels is greater than the number of corrupted one. In the given example, $\#$ $uncorrupted$ $pixel=3$ while $\#$ $corrupted$ $pixel=4$, which followed by ignoring this window and enlarge it to be $5\times5$ window.
\item In $5\times5$ window, $\#$ $uncorrupted$ $pixel=14$ while $\#$ $corrupted$ $pixel=11$. This window considered accepted window, and the threshold value will be obtained by taking the median value of the uncorrupted pixels, $T = med$\big\{${255,47,255,50,0,0,224,62,0,255,255,72,0,179}$\big\} and equal to $67$.
\end{itemize}

\begin{algorithm}
  \KwData{The original image $I$, the image of labeled pixels $C$, maximum window size $W_m$.}
  \KwResult{Pixels threshold values $T$, and pixels corresponding window size $WS$.}
\For{each pixel position \boldsymbol{$x_0$}}{ 
\uIf{$I(x_0)$ is classified as $uncorrupted$ $pixel$ in $C$}{
$T(x_0) = I(x_0)$\\
$WS(x_0) = 1$\\
}
 \Else{
Initialize $w=3$\\
Impose a window $W$ ($w \times w$) around $x_0$ ($W$ $\in$ $I$)\\
Intialize $WS(x_0) = W_m$\\
\While{$W<W_m$}{
Find $N_{un}=\#$ $uncorrupted$ $pixels$ in $W$\\
\uIf{$N_{un} \geq \frac{W^{2}}{2}$}{
$WS(x_0)=W$\\
Break\\
}
\Else{
Update $W$($w \times w$), where $w=w+2$\\
}}
Find $I_{un}$ $(uncorrupted$ $pixels$ $in$ $WS(x_0))$\\
$T(x_0) = med(I_{un})$\\
}}
  \caption{Generation of local thresholds}
  \label{Alg2}
\end{algorithm}

\subsection{Generate the binary pattern}
\label{sec:BM}

\begin{figure}[t]
\centering
\includegraphics[width=3.5in]{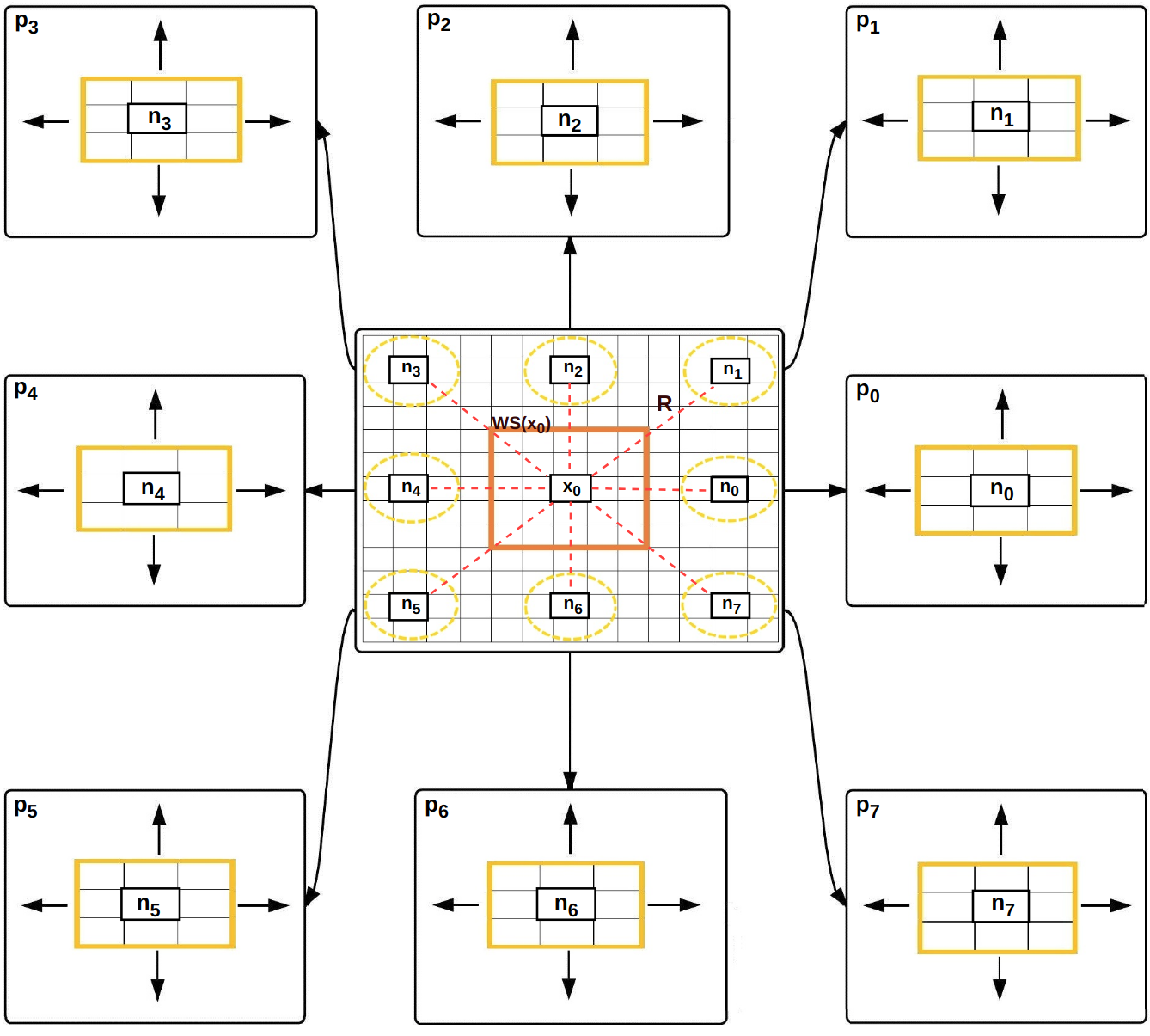}
\DeclareGraphicsExtensions.
\caption{Binary module scheme (Alg.~\ref{Alg3}). Where $x_0$ is the current central pixel, $WS(x_0)$ is $x_0$ corresponding window size (Alg.~\ref{Alg2}), and $P$ are the neighborhood patches with each central pixel ($n$).}
\label{fig 4}
\end{figure}

To reach the highest performance in texture classification, the descriptor should balance the classification goals such as robustness to noise, discriminativeness, and low computational cost. LBP descriptor conveys local structures, but to achieve better performance, discriminative properties should be used by considering the effect of image patches instead of taking a single pixel. To provide more information to the descriptor, these patches do not intersect with central pixel threshold window (Sec~\ref{sec:TP}). As well as, and each patch size will be found using an adaptive way that depends on each patch pixels. Fig.~\ref{fig 4} and Alg.~\ref{Alg3} demonstrates the binary pattern module of the proposed descriptor.

The binary pattern module (Alg.~\ref{Alg3}) represents the procedure of forming the binary pattern. The module starts by finding the neighborhood patches with a maximum size around its central pixel. For each patch, a $3\times3$ window imposed around its central pixel. If the number of uncorrupted pixels is more than the corrupted pixels, this window considered accepted window and the value of the patch is the median of the uncorrupted pixels in that window. Otherwise, the window is enlarged to be $5\times5$ window. The process continues until reaching the predefined maximum window size. After finding each neighborhood patch value, the binary pattern ($8bits$) is computed with a simple procedure between the patches values and the central pixel threshold value, where each patch represented in the binary pattern by 0 or 1. 

\begin{algorithm}
  \KwData{The original image $I$, the image of labeled pixels $C$, maximum window size $W_m$, pixels threshold values $T$, and pixels corresponding window size $WS$.}
  \KwResult{The binary pattern (RAMBP).}
\For{each pixel position \boldsymbol{$x_0$}}{ 
The distance between the central pixel and each patch center ($n_i$): $R =$ $W_m$ + $WS(x_0)$\\
\For{each patch $P_i$ ($i\in{0:7}$)}{ 
Initialize $w=3$\\
Impose a window $W$ ($w \times w$) around $P_i$ center ($n_i$) ($W$ $\in$ $I$)\\
Intialize patch window size $W_{P_i} = W_m$\\
\While{$W<W_m$}{
Find $N_{un}=\#$ $uncorrupted$ $pixels$ in $W$\\
\uIf{$N_{un} \geq \frac{W^{2}}{2}$}{
$W_{P_i}=W$\\
Break.
}
\Else{
Update $W$($w \times w$), where $w=w+2$\\
}}
Find $I_{un}$ $(uncorrupted$ $pixels$ $in$ $W_{P_i})$\\
$\beta = med(I_{un})$

$ S(P_i)= \begin{cases}
    1 , &\quad  T(x_0) \geq \beta \\
    0 , & \quad T(x_0) < \beta\\
  \end{cases}$
}
$\ The$ $binary$ $pattern(x_0)= \sum_{i = 0}^{7}{S(P_i) 2^i}$
}
  \caption{Generation of binary patterns}
  \label{Alg3}
\end{algorithm}

\begin{table*}[htbp]
  \caption{Summary of the used Datasets.}
\makebox[\textwidth][l]{
    \begin{tabular}{lcccl}
\toprule
Texture datasets & $\#$ of classes & $\#$ of images & Image size($pxls$) & Challenges\\ 
\midrule
$Outex\_TC10$~\cite{ojala2002outex}&24&4320&$128\times128$&Rotation changes \\
$Outex\_TC11$~\cite{ojala2002outex}&24&960&$128\times128$&Inca illuminant, rotations ($0^{\circ}$)\\
$Outex\_TC12$~\cite{ojala2002outex}&24&4800&$128\times128$&Illumination variations, rotation changes\\
$Outex\_TC23$~\cite{ojala2002outex}&68&2720&$128\times128$&Inca illuminant, rotations ($0^{\circ}$)\\
$Curet$~\cite{varma2005statistical}&61&5612&$200\times200$&Illumination variation, rotations and pose changes, specularities, shadowing\\
$Brodatz$~\cite{brodatz1966textures}&111&999&$215\times215$&Large number of classes, lack of intraclass variations\\
$BrodatzRot$~\cite{brodatz1966textures}&111&999&$128\times128$&Rotation changes, large number of classes, lack of intraclass variations\\
$KTH-TIPS2b$~\cite{mallikarjuna2006kth}&11&4752&$200\times200$&Pose changes, illumination changes, scale changes\\
$ALOT$~\cite{burghouts2009material}&250&25000&$384\times256$&Strong illumination changes, rotation changes, large number of classes\\
\bottomrule
\end{tabular}%
}
\label{tab1}%
\end{table*}%

\begin{figure*}[htbp]
\rotatebox[origin=c]{90}{\bfseries $Outex\_TC11$\strut}
	\begin{subfigure}{0.15\textwidth}
     \includegraphics[width=\linewidth]{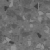}
        \caption{$Noise-free$}
    \end{subfigure}
        \begin{subfigure}{0.15\textwidth}
     \includegraphics[width=\linewidth]{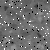}
        \caption{$\rho= 5\%$}
    \end{subfigure}
    \begin{subfigure}{0.15\textwidth}
     \includegraphics[width=\linewidth]{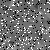}
        \caption{$\rho= 15\%$}
    \end{subfigure}
    \begin{subfigure}{0.15\textwidth}
     \includegraphics[width=\linewidth]{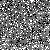}
        \caption{$\rho= 30\%$}
    \end{subfigure}
    \begin{subfigure}{0.15\textwidth}
     \includegraphics[width=\linewidth]{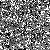}
        \caption{$\rho= 40\%$}
    \end{subfigure}
        \begin{subfigure}{0.15\textwidth}
     \includegraphics[width=\linewidth]{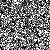}
        \caption{$\rho= 50\%$}
    \end{subfigure}
    
\rotatebox[origin=c]{90}{\bfseries $Outex\_TC23$\strut}
	\begin{subfigure}{0.15\textwidth}
     \includegraphics[width=\linewidth]{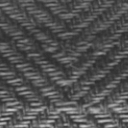}
        \caption{$Noise-free$}
    \end{subfigure}
               \begin{subfigure}{0.15\textwidth}
        \includegraphics[width=\linewidth]{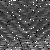}
        \caption{$\rho= 5\%$}
    \end{subfigure}
            \begin{subfigure}{0.15\textwidth}
        \includegraphics[width=\linewidth]{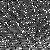}
        \caption{$\rho= 15\%$}
    \end{subfigure}
            \begin{subfigure}{0.15\textwidth}
        \includegraphics[width=\linewidth]{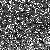}
        \caption{$\rho= 30\%$}
    \end{subfigure}
        \begin{subfigure}{0.15\textwidth}
        \includegraphics[width=\linewidth]{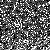}
        \caption{$\rho= 40\%$}
    \end{subfigure}
        \begin{subfigure}{0.15\textwidth}
        \includegraphics[width=\linewidth]{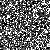}
        \caption{$\rho= 50\%$}
    \end{subfigure}
    \caption{Example of $Outex\_TC11$ and $Outex\_TC23$ textures with different impulse noise densities.}
    \label{fig2}
\end{figure*}

\section{Experiments and results}
\label{sec:EX}

The experiments were carried out with a core 7 Duo 3.50 GHz processor with 32GB RAM under Matlab. Nine texture datasets were conducted in these experiments, which considered from the most commonly used texture datasets. Tab.~\ref{tab1} summarized the used texture datasets, number of classes, number of images, images size, and each texture challenges. 

To evaluate the robustness of the proposed approach, $k$-nearest neighbor ($k$-NN) had been used. The $k$-NN classifier recognized as one of the most popular and simplest methods, the $k$-NN is used with $\chi^2$ distance defined as

\begin{equation}‎\label{eq:1}
‎‎\ \chi^2(x,y)‎ ‎=‎ \frac{1}{2}\mathlarger{‎‎\sum}_{i}{}{\frac{(x_i-y_i)^2}{x_i+y_i}}
\end{equation}‎
 
where $x$ and $y$ are the features vectors of two different textures. $k$-NN is adopted with $k$ value equal to $1$ for most experiments, but this parameter has been varied to test its influence on the performance consistency.

\begin{figure}[t]
\centering
\includegraphics[width=2.5in]{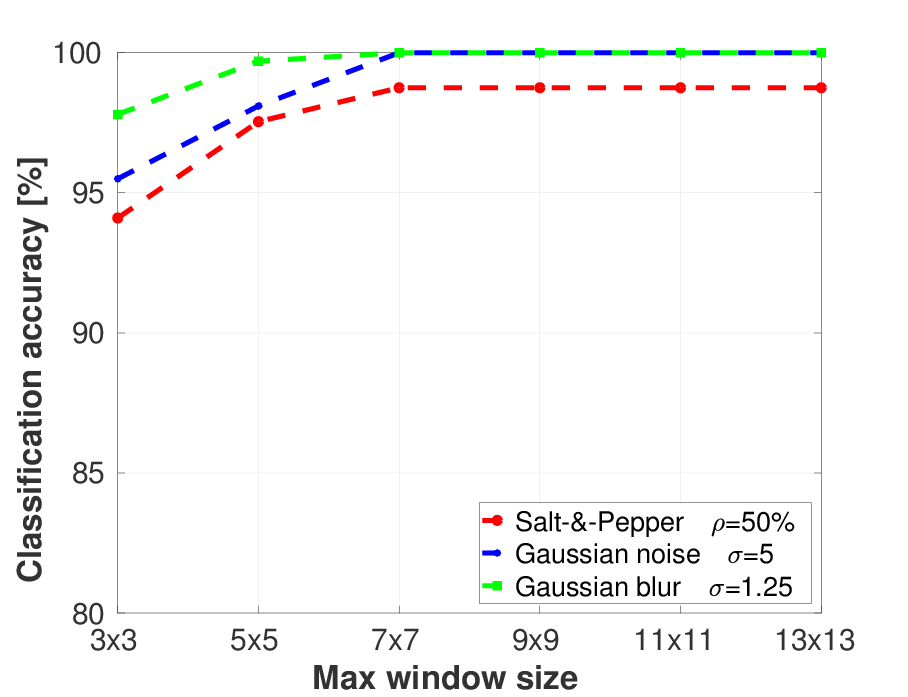}
\DeclareGraphicsExtensions.
\caption{Illustration of the performance according to the maximum window size that the adaptive window could reach.}
\label{fig3}
\end{figure}

In order to study the effect of the adaptive window maximum size, RAMBP performance has been tested on $Outex\_TC11$ dataset with different maximum window sizes. Fig.~\ref{fig3} shows the classification score in different applied noises for different maximum window size values. As can be seen, the larger window size gives the higher score, but the time complexity will grow exponentially. Therefore, a good trade-off should be taken between the accuracy and the time complexity. In the experiments, $5\times5$ max window size is adopted since it gives high classification score and makes the algorithm run faster. In comparison with traditional LBP, the proposed method is slower but it has less computational complexity and dimensionality than many LBP descriptors used to address the noisy textures.

In this section, we start by evaluation of the proposed method on high noisy textures, including Salt-and-Pepper noise, Gaussian noise, and Gaussian blur. Followed by experiment results of the proposed method on noise-free textures. Finally, the evaluation of the proposed method for noisy textures retrieval. In this paper, some of state-of-the-art descriptors results have been reported from~\cite{liu2017local}.

\subsection{Noisy texture classification}

Noise robustness is a crucial point for evaluating descriptors. In this experiment, in order to test the noisy textures and evaluate the descriptor robustness in a more accurate way, the random noise generation has been repeated $10$ times over the dataset, and the classification results had been noted by taking the average of these $10$ tests.  Noise-free images have been used for the training step while testing step performed on the noisy images. Choosing this scheme makes the noisy texture classification very difficult since the descriptor does not use any noise information and any prior learning process.

\begin{table*}[htbp]
  \caption{Classification scores ($\%$) comparison between the proposed descriptor (RAMBP) and state-of-the-art descriptors for Salt-and-Pepper noise.}
\makebox[\textwidth][c]{
    \begin{tabular}{lcccccccccc}
\toprule
Dataset & \multicolumn{5}{c}{$Outex\_TC11$} & \multicolumn{5}{c}{$Outex\_TC23$} \\
Noise parameter & \multicolumn{5}{c}{Noise density $\rho$} & \multicolumn{5}{c}{Noise density $\rho$} \\
\cmidrule(lr){1-1}
\cmidrule(lr){2-6}
\cmidrule(lr){7-11}
Method & $5\%$ & $15\%$ & $30\%$ & $40\%$  & $50\%$ & $5\%$ & $15\%$ & $30\%$ & $40\%$  & $50\%$\\ 
\midrule
$LBP$~\cite{ojala2002multiresolution} &85.4&15.5&5.4&4.2&4.2&66.0&9.9&3.8&1.8&1.5\\
$LBPriu2$~\cite{ojala2002multiresolution} &31.7&4.2&4.2&4.4&4.2&11.8&1.5&1.5&1.5&1.5\\
$LBPri$~\cite{ojala2002multiresolution} &47.1&10.0&4.2&4.2&4.2&26.5&4.7&2.2&1.5&1.5\\
$ILBPrui2$~\cite{jin2004face} &27.3&4.2&4.2&4.2&4.2&10.7&2.1&1.5&1.5&1.5\\
$CLBP$~\cite{guo2010completed}&17.3&8.3&4.2&4.2&4.2&7.6&2.9&1.5&1.6&1.5\\
$MBPriu2$~\cite{hafiane2007median} &31.0&8.3&4.2&4.2&4.2&17.0&2.5&1.5&1.5&1.5\\
$MBP$~\cite{hafiane2007median}&95.8&38.6&20.5&16.6&16.1&76.8&18.6&6.0&4.9&4.2\\
$RLBPriu2$~\cite{chen2013rlbp} &39.2&4.2&4.2&4.2&4.2&18.5&1.5&1.5&1.5&1.5\\
$EXLBP$~\cite{zhou2008novel} &27.3&4.2&4.2&4.2&4.2&12.2&1.5&1.5&1.5&1.5\\
$NTLBP$~\cite{fathi2012noise} &74.4&22.1&4.8&5.0&6.3&40.5&4.7&3.8&2.6&2.7\\
$MDLBPriu2$~\cite{schaefer2012multi} &71.9&13.5&8.3&4.2&4.2&38.2&3.7&2.9&2.5&1.9\\
$DLBP$~\cite{liao2009dominant} &29.8&5.4&4.2&4.2&4.2&16.5&4.9&1.5&1.5&1.5\\
$BRINT$~\cite{liu2014brint} &30.8&7.1&6.0&4.4&4.2&15.9&1.5&1.5&1.3&1.5\\
$LBPD$~\cite{hong2014combining} &25.2&8.3&4.2&4.2&4.2&10.3&2.9&1.5&1.5&0.1\\
$SSLBP$~\cite{guo2016robust} &29.0&9.6&4.2&4.2&4.2&24.5&2.8&1.5&1.5&1.5\\
$AMBP$~\cite{hafiane2015joint} &100.0&95.4&20.7&13.8&10.7&100.0&85.0&4.8&1.8&1.5\\
$MRELBP$~\cite{liu2016median} &100.0&100.0&100.0&85.8&50.2&100.0&99.9&94.0&54.6&19.2\\
$FV-VGGVD(SVM)$~\cite{cimpoi2015deep}&21.0&12.1&6.0&6.5&4.2&10.3&5.2&2.3&1.5&1.8\\
$RAMBP$ &\textbf{100.0}&\textbf{100.0}&\textbf{100.0}&\textbf{99.1}&\textbf{98.5}&\textbf{100.0}&\textbf{100.0}&\textbf{100.0}&\textbf{99.8}&\textbf{90.2}\\

\bottomrule
\end{tabular}%
}
\label{tab2}%
\end{table*}%

\begin{figure*}[htbp]
    \centering
    \begin{subfigure}[b]{0.3\textwidth}
        \includegraphics[width=\textwidth]{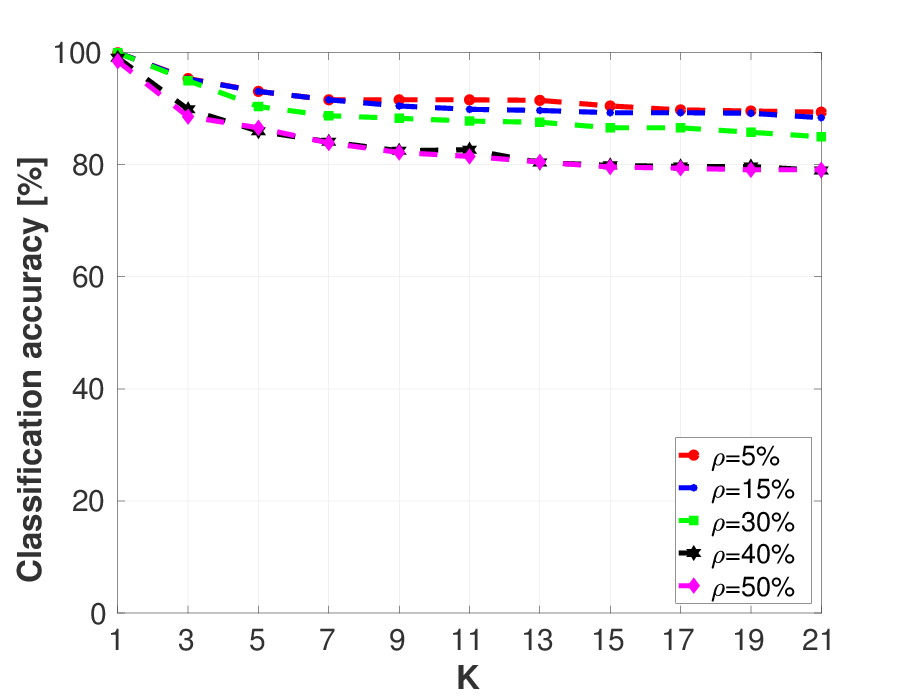}
        \caption{$Outex\_TC11$}
        \label{sp11}
    \end{subfigure}
    \begin{subfigure}[b]{0.3\textwidth}
        \includegraphics[width=\textwidth]{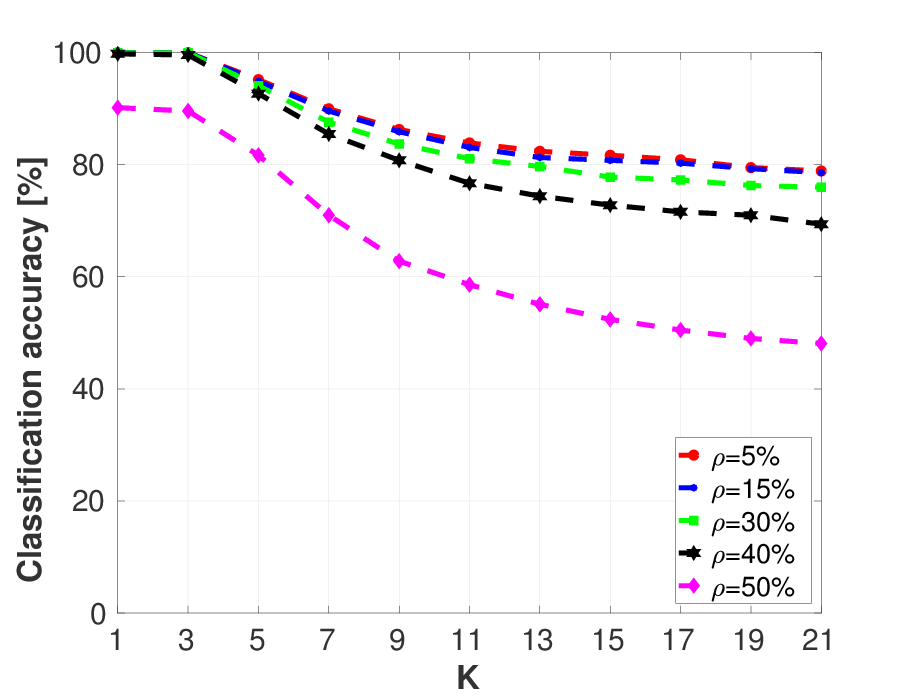}
        \caption{$Outex\_TC23$}
        \label{sp23}
    \end{subfigure}
    \caption{The performance of RAMBP for Salt-and-Pepper noise according to $k$ values in $k$-NN. Where $k$ starts from $1$ to $21$ with a step of two to avoid tie problem, and $\rho$ parameter indicates Salt-and-Pepper density.}
    \label{sp}
\end{figure*}

\subsubsection{Salt-and-Pepper noise}
\label{sec:SP}
Impulse noise introduces high or low values randomly distributed over the image. Salt-and-Pepper noise has been applied to $Outex\_TC11$ and $Outex\_TC23$ datasets with different noise densities $\rho$. High noisy textures are very challenging as it can be seen in Fig.~\ref{fig2} where textures are visually unrecognizable from $30\%$ of noise. 

The results of the proposed algorithm are listed in Tab.~\ref{tab2}. It can be observed that the classification accuracy is improved after using the proposed method. Compared to the different state-of-the-art techniques, RAMBP yields the best results and outperforms other techniques, especially on high noisy textures.

As can be seen from the results, using rotational uniform scheme decreases the performance of the LBP based descriptors. Using $LBP$ and $MBP$ gave better results than $LBPriu2$ and $MBPriu2$, respectively. It can also be noticed from Tab.~\ref{tab2} that, $MRELBP$ offers the second best performance but its accuracy drops drastically with high noise densities (e.g. $50\%$). Also, $AMBP$ gives good results and noise robustness under low-density impulse noise but not for high noise.

Although RAMBP previously mentioned performance shows a high score where KNN ($k=1$) provides the best match among all images, it is important to study the matched percentage of the same class images. This percentage can be computed using different $k$ values in $k$-NN. In other words, an image is classified by the majority votes and assigned to the most common class. For example $k=1$, KNN provides the nearest image, then the examined image will be classified as that image class. For $k=3$, KNN provides the nearest three images and the examined image will be classified to class with the majority votes between the three images classes. For that, RAMBP performance has been tested with different $k$ values in $k$-NN. We can notice from Fig.~\ref{sp11} the stability and robustness of RAMBP in different Salt-and-Pepper noise densities where it keeps good accuracy even with high noise density and large value of $k$. Also shown in Fig.~\ref{sp23}, the descriptor performance over $k$ has a more decreasing rate which is proportional to the noise density. This happens may be due to the number of classes (i.e. 68) in $Outex\_TC23$ dataset. Nevertheless, the accuracy stays good for different $k$ values. 

\begin{table*}[htbp]
\caption{Classification scores ($\%$) comparison between the proposed descriptor (RAMBP) and state-of-the-art descriptors for Gaussian noise with standard deviation $\sigma$.}
\makebox[\textwidth][c]{
    \begin{tabular}{lcc}
\toprule
Dataset & \multicolumn{1}{c}{$Outex\_TC11$} & \multicolumn{1}{c}{$Outex\_TC23$} \\
\cmidrule(lr){1-1}
\cmidrule(lr){2-2}
\cmidrule(lr){3-3}
Method & $\sigma = 5$ & $\sigma = 5$ \\ 
\midrule
$LBP$~\cite{ojala2002multiresolution}&35.0&09.8\\
$LBPriu2$~\cite{ojala2002multiresolution}&17.7&8.4\\
$LBPri$~\cite{ojala2002multiresolution}&16.0&7.9\\
$ILBPrui2$~\cite{jin2004face}&17.5&10.4\\
$CLBP$~\cite{guo2010completed}&11.9&5.6\\
$MBPriu2$~\cite{hafiane2007median}&12.1&5.2\\
$MBP$~\cite{hafiane2007median}&59.4&22.0\\
$RLBPriu2$~\cite{chen2013rlbp}&22.1&11.9\\
$EXLBP$~\cite{zhou2008novel}&19.2&10.3\\
$NTLBP$~\cite{fathi2012noise}&24.0&9.0\\
$MDLBPriu2$~\cite{schaefer2012multi}&12.5&6.1\\
$DLBP$~\cite{liao2009dominant}&14.8&8.2\\
$BRINT$~\cite{liu2014brint}&61.9&27.4\\
$LBPD$~\cite{hong2014combining}&24.6&14.8\\
$SSLBP$~\cite{guo2016robust}&97.1&91.5\\
$AMBP$~\cite{hafiane2015joint}&96.5&74.3\\
$MRELBP$~\cite{liu2016median}&91.5&79.2\\
$FV-VGGVD(SVM)$~\cite{cimpoi2015deep}&93.1&71.5\\
$RAMBP$ &\textbf{99.0}&\textbf{95.9}\\

\bottomrule
\end{tabular}%
}
\label{tab3}%
\end{table*}%

\begin{figure*}[htbp]
    \centering
    \begin{subfigure}[b]{0.26\textwidth}
        \includegraphics[width=\textwidth]{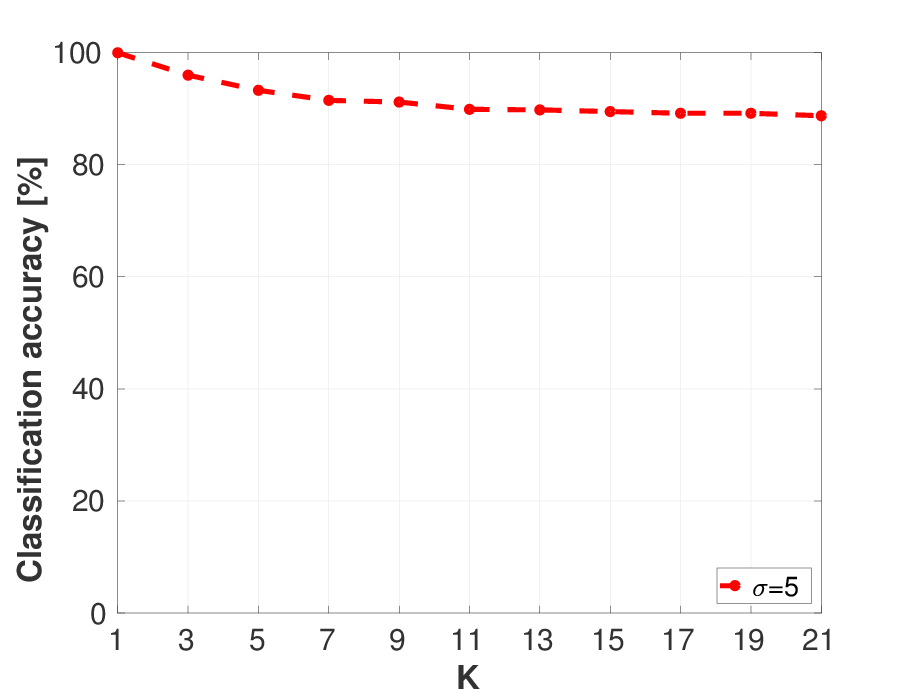}
        \caption{$Outex\_TC11$}
    \end{subfigure}
    \begin{subfigure}[b]{0.26\textwidth}
        \includegraphics[width=\textwidth]{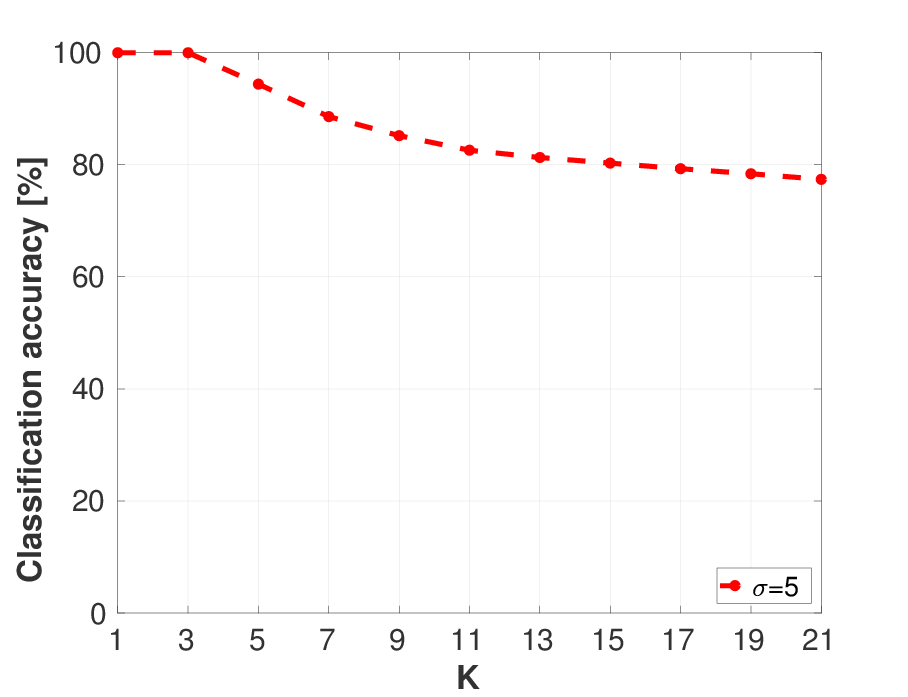}
        \caption{$Outex\_TC23$}
    \end{subfigure}
 \caption{The performance of RAMBP for Gaussian noise with different $k$ values in $k$-NN. Where $k$ starts from $1$ to $21$ with a step of two to avoid tie problem, and $\sigma$ parameter indicates Gaussian noise standard deviation.}
     \label{gn}
\end{figure*}

\begin{figure}[htbp]
\center
\rotatebox[origin=c]{90}{\small  $Outex\_TC11$}
	\begin{subfigure}{0.14\textwidth}
     \includegraphics[width=\linewidth]{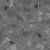}
        \caption{$Noise-free$}
    \end{subfigure}
        \begin{subfigure}{0.14\textwidth}
     \includegraphics[width=\linewidth]{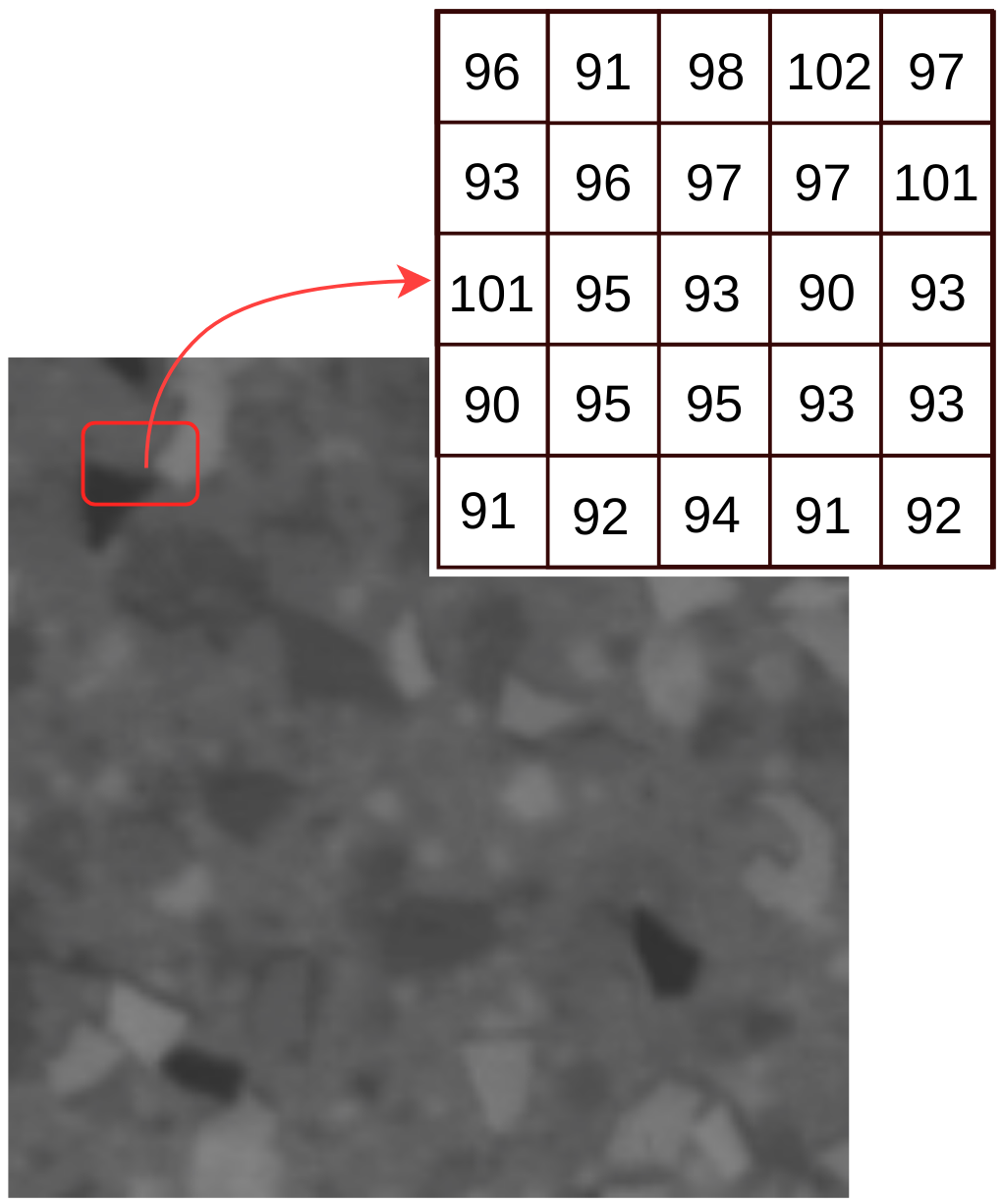}
        \caption{$\sigma= 5$}
    \end{subfigure}

\rotatebox[origin=c]{90}{\small  $Outex\_TC23$}
	\begin{subfigure}{0.14\textwidth}
     \includegraphics[width=\linewidth]{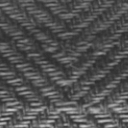}
        \caption{$Noise-free$}
    \end{subfigure}
               \begin{subfigure}{0.14\textwidth}
        \includegraphics[width=\linewidth]{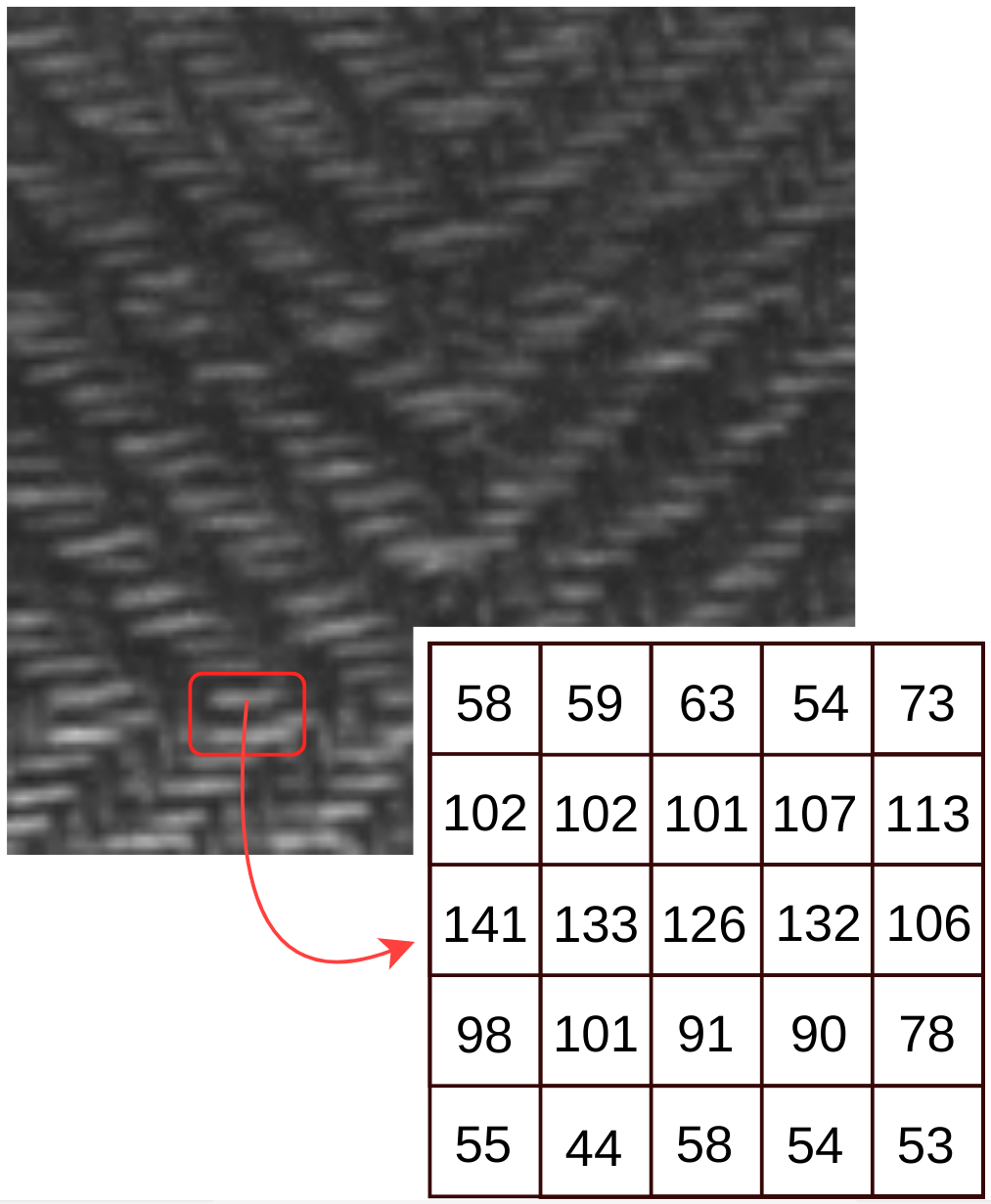}
        \caption{$\sigma= 5$}
    \end{subfigure}
 \caption{Example of $Outex\_TC11$ and $Outex\_TC23$ textures with Gaussian noise standard deviation $\sigma= 5$, where it shows the changes in pixels values. }
    \label{fig5}
\end{figure}

\subsubsection{Gaussian noise}
\label{sec:GN}
Gaussian noise is an additive noise affects digital images gray values. Gaussian noise has been added to $Outex\_TC11$ and $Outex\_TC23$ datasets with standard deviation $\sigma = 5$. Fig.~\ref{fig5} provides an example of the used datasets after adding Gaussian noise, where visually it is difficult to see the global effect and the difference between noise-free and noisy textures, but it can be seen the local information and pixels intensity are affected.

Tab.~\ref{tab3} shows the classification results of the proposed method as well as the state-of-the-art descriptors, where RAMBP provides the best performance among other descriptors. SSLBP descriptor gives the second best results, followed by MRELBP, AMBP, and deep learning techniques. But SSLBP yielded poor accuracy under Salt-and-Pepper as indicated in Tab.~\ref{tab2}. As can be observed from Tab.~\ref{tab2} and Tab.~\ref{tab3}, the proposed method achieved the best results in both experiments and showed nice consistency in different types of noise. 

To illustrate the stability of RAMBP for Gaussian noise, different $k$ values in $k$-NN have been tested as shown in Fig.~\ref{gn}. We can notice, a small decrease of RAMBP accuracy with increasing the value of $k$. Overall, RAMBP provides good stability and robustness even at large values of $k$ in $k$-NN.

\subsubsection{Gaussian blur}
\label{sec:GB}

\begin{figure*}[htbp]
\center
\rotatebox[origin=c]{90}{\bfseries $Outex\_TC11$\strut}
	\begin{subfigure}{0.15\textwidth}
     \includegraphics[width=\linewidth]{TC11-noise-free}
        \caption{$Noise-free$}
    \end{subfigure}
        \begin{subfigure}{0.15\textwidth}
     \includegraphics[width=\linewidth]{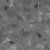}
        \caption{$\sigma= 0.5$}
    \end{subfigure}
    \begin{subfigure}{0.15\textwidth}
     \includegraphics[width=\linewidth]{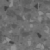}
        \caption{$\sigma= 0.75$}
    \end{subfigure}
    \begin{subfigure}{0.15\textwidth}
     \includegraphics[width=\linewidth]{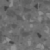}
        \caption{$\sigma= 1$}
    \end{subfigure}
    \begin{subfigure}{0.15\textwidth}
     \includegraphics[width=\linewidth]{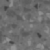}
        \caption{$\sigma= 1.25$}
    \end{subfigure}
    
\rotatebox[origin=c]{90}{\bfseries $Outex\_TC23$\strut}
	\begin{subfigure}{0.15\textwidth}
     \includegraphics[width=\linewidth]{TC23-noise-free}
        \caption{$Noise-free$}
    \end{subfigure}
               \begin{subfigure}{0.15\textwidth}
        \includegraphics[width=\linewidth]{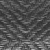}
        \caption{$\sigma= 0.5$}
    \end{subfigure}
            \begin{subfigure}{0.15\textwidth}
        \includegraphics[width=\linewidth]{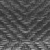}
        \caption{$\sigma= 0.75$}
    \end{subfigure}
            \begin{subfigure}{0.15\textwidth}
        \includegraphics[width=\linewidth]{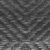}
        \caption{$\sigma= 1$}
    \end{subfigure}
        \begin{subfigure}{0.15\textwidth}
        \includegraphics[width=\linewidth]{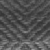}
        \caption{$\sigma= 1.25$}
    \end{subfigure}
    \caption{Example of $Outex\_TC11$ and $Outex\_TC23$ textures with different Gaussian blur standard deviation.}
    \label{fig7}
\end{figure*}

Gaussian blur, known also as Gaussian smoothing, is another kind of effects happened to images, which results in removing image detail. This type of noise also modify the local structure which affects the local binary patterns. In these experiments, Gaussian blur has been applied to $Outex\_TC11$ and $Outex\_TC23$ datasets with different standard deviations $\sigma$. Fig.~\ref{fig7} illustrates an example of the used datasets with Gaussian blur.

\begin{table*}[htbp]
  \caption{Classification scores ($\%$) comparison between the proposed descriptor (RAMBP) and state-of-the-art descriptors for Gaussian blur with standard deviation $\sigma$.}
\makebox[\textwidth][c]{
    \begin{tabular}{lcccccccc}
\toprule
Dataset & \multicolumn{4}{c}{$Outex\_TC11$} & \multicolumn{4}{c}{$Outex\_TC23$} \\
\cmidrule(lr){1-1}
\cmidrule(lr){2-5}
\cmidrule(lr){6-9}
Method & $\sigma = 0.5$ & $\sigma = 0.75$ & $\sigma = 1$ & $\sigma = 1.25$ & $\sigma = 0.5$ & $\sigma = 0.75$ & $\sigma = 1$ & $\sigma = 1.25$ \\  
\midrule
$LBP$~\cite{ojala2002multiresolution}&99.1&71.8&53.8&39.8&99.9&55.0&37.8&29.0\\
$LBPriu2$~\cite{ojala2002multiresolution}&94.2&46.5&24.6&12.7&72.4&30.3&16.6&9.7\\
$LBPri$~\cite{ojala2002multiresolution}&86.9&44.6&26.0&18.1&57.7&28.3&16.0&9.4\\
$ILBPrui2$~\cite{jin2004face}&97.3&59.8&29.4&20.4&81.7&43.2&25.1&16.7\\
$CLBP$~\cite{guo2010completed}&98.8&74.8&49.6&23.1&86.6&55.4&36.1&21.2\\
$MBPriu2$~\cite{hafiane2007median}&85.4&29.0&18.5&11.9&58.7&22.5&13.5&10.6\\
$MBP$~\cite{hafiane2007median}&97.5&73.0&50.9&39.5&99.0&58.2&40.7&28.3\\
$RLBPriu2$~\cite{chen2013rlbp}&95.0&49.8&28.7&16.5&75.4&33.2&18.4&10.7\\
$EXLBP$~\cite{zhou2008novel}&94.0&47.7&28.3&17.1&73.3&32.0&17.8&10.5\\
$NTLBP$~\cite{fathi2012noise}&96.3&49.0&33.1&19.4&80.1&35.7&21.7&14.1\\
$MDLBPriu2$~\cite{schaefer2012multi}&100.0&60.2&36.9&23.8&95.7&35.1&20.6&12.2\\
$DLBP$~\cite{liao2009dominant}&90.4&61.5&21.9&13.1&67.7&31.3&16.5&8.7\\
$BRINT$~\cite{liu2014brint}&100.0&97.1&80.4&44.6&100&97.5&59.1&39.1\\
$LBPD$~\cite{hong2014combining}&99.4&85.8&65.2&45.4&87.7&56.0&40.2&30.6\\
$SSLBP$~\cite{guo2016robust}&\textbf{100.0}&\textbf{100.0}&\textbf{100.0}&\textbf{100.0}&\textbf{100.0}&\textbf{100.0}&\textbf{100.0}&90.6\\
$AMBP$~\cite{hafiane2015joint}&100.0&99.0&88.7&52.6&100.0&99.5&81.4&53.8\\
$MRELBP$~\cite{liu2016median}&100.0&100.0&93.8&75.4&99.9&97.9&85.8&61.8\\
$FV-VGGVD(SVM)$~\cite{cimpoi2015deep}&100.0&100.0&96.5&89.8&99.6&94.1&83.1&71.8\\
$RAMBP$ &\textbf{100.0}&\textbf{100.0}&\textbf{100.0}&\textbf{100.0}&\textbf{100.0}&\textbf{100.0}&\textbf{100.0}&\textbf{99.2}\\

\bottomrule
\end{tabular}%
}
\label{tab4}%
\end{table*}%

\begin{figure*}[htbp]
    \centering
    \begin{subfigure}[b]{0.3\textwidth}
        \includegraphics[width=\textwidth]{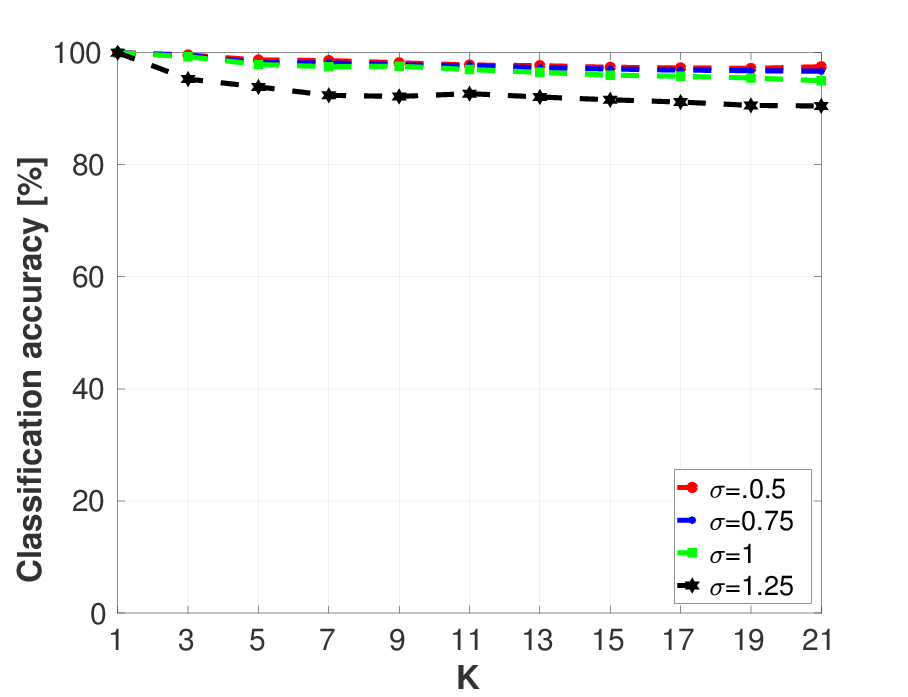}
        \caption{$Outex\_TC11$}
    \end{subfigure}
    \begin{subfigure}[b]{0.3\textwidth}
        \includegraphics[width=\textwidth]{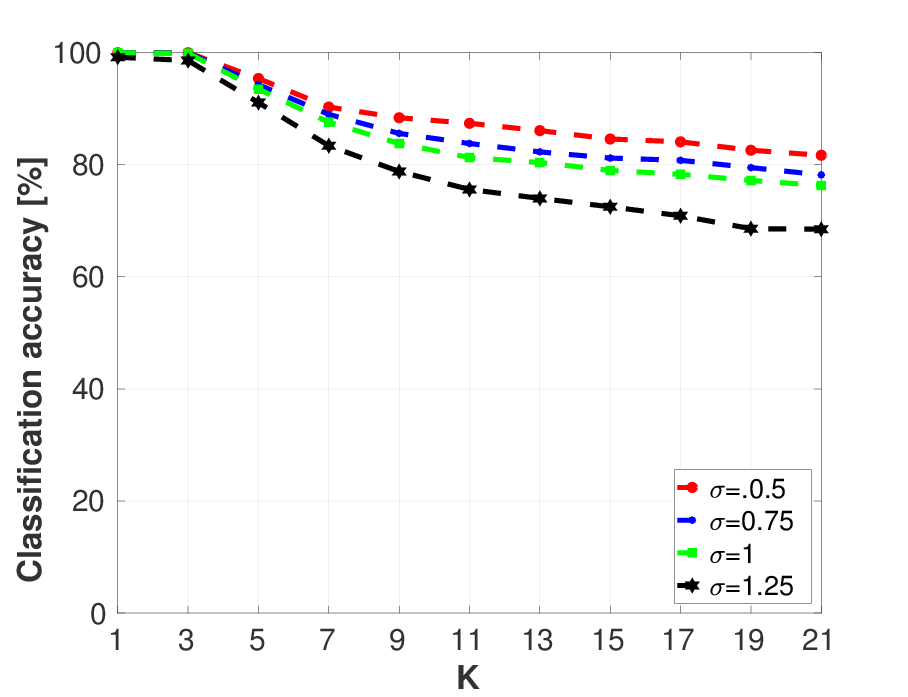}
        \caption{$Outex\_TC23$}
    \end{subfigure}
    \caption{The performance of RAMBP for Gaussian blur with different $k$ values in $k$-NN. Where $k$ starts from $1$ to $21$ with a step of two to avoid tie problem, and $\sigma$ parameter indicates Gaussian blur standard deviation.}
    \label{gb}
\end{figure*}

\begin{table*}[htbp]
  \caption{Classification scores ($\%$) comparison between the proposed descriptor (RAMBP) and state-of-the-art descriptors for noise-free texture classification.}
\makebox[\textwidth][c]{
    \begin{tabular}{lccccccc}
\toprule
Method & $TC10$ & $TC12$ & $Curet$ & $Brodatz$ & $BrodatzRot$ & $KTH2b$ & $ALOT$\\ 
\cmidrule(lr){1-1}
\cmidrule(lr){2-2}
\cmidrule(lr){3-3}
\cmidrule(lr){4-4}
\cmidrule(lr){5-5}
\cmidrule(lr){6-6}
\cmidrule(lr){7-7}
\cmidrule(lr){8-8}
$\# classes$ & $(24)$ & $(24)$ & $(61)$ & $(111)$ & $(111)$ & $(11)$ & $(250)$\\  
\midrule
$LBP$~\cite{ojala2002multiresolution}&99.36&90.55&92.77&88.67&76.48&60.33&86.58\\
$LBPriu2$~\cite{ojala2002multiresolution}&99.69&92.16&97.03&90.70&79.22&62.69&94.15\\
$LBPri$~\cite{ojala2002multiresolution}&86.69&83.68&95.38&89.93&71.73&61.48&93.29\\
$ILBPrui2$~\cite{jin2004face}&99.66&93.34&94.66&91.66&82.27&61.93&95.71\\
$CLBP$~\cite{guo2010completed}&99.45&95.78&97.33&92.34&84.35&64.18&96.74\\
$MBPriu2$~\cite{hafiane2007median}&95.29&86.69&92.09&87.25&74.57&61.49&88.23\\
$MBP$~\cite{hafiane2007median}&98.52&97.17&91.24&89.27&76.67&60.19&91.30\\
$RLBPriu2$~\cite{chen2013rlbp}&99.66&93.53&97.20&91.09&79.59&61.20&94.23\\
$EXLBP$~\cite{zhou2008novel}&99.64&93.55&96.85&90.19&80.08&62.39&95.20\\
$NTLBP$~\cite{fathi2012noise}&99.32&95.27&96.11&89.31&80.25&61.30&94.47\\
$MDLBPriu2$~\cite{schaefer2012multi}&99.22&95.64&96.92&93.40&82.31&66.52&95.81\\
$DLBP$~\cite{liao2009dominant}&99.46&91.97&94.38&88.73&75.04&61.72&NO\\
$BRINT$~\cite{liu2014brint}&99.35&98.13&97.02&90.83&78.77&66.67&96.13\\
$LBPD$~\cite{hong2014combining}&98.78&96.67&94.23&89.74&74.79&63.47&92.82\\
$SSLBP$~\cite{guo2016robust}&99.82&99.36&98.79&89.94&80.03&65.57&96.68\\
$AMBP$~\cite{hafiane2015joint}&99.68&98.12&95.64&90.67&79.86&62.73&95.82\\
$MRELBP$~\cite{liu2016median}&99.82&99.58&97.10&90.86&81.92&\textbf{68.98}&97.28\\
$FV-VGGVD(SVM)$~\cite{cimpoi2015deep}&80.00&82.30&\textbf{99.00}&\textbf{98.70}&\textbf{92.10}&\textbf{88.20}&\textbf{99.50}\\
$RAMBP$&\textbf{99.90}&\textbf{99.70}&\textbf{98.50}&\textbf{94.05}&\textbf{86.98}&68,86&\textbf{97.59} \\

\bottomrule
\end{tabular}%
}
\label{tab5}%
\end{table*}%

Tab.~\ref{tab4} depicts the classification scores after applying Gaussian blur. The proposed method shows the best score with SSLBP method. The latest method performs nicely here because it includes the blurring process in descriptor generation. However, it must be recalled the poor performance of SSLBP in Salt-and-Pepper noise as evidenced in Tab.~\ref{tab2}. MRELBP and FV-CNN have good performance under low noisy textures, but the accuracy vastly decreases under higher noise.

The accuracy of the proposed method can also be observed in Fig.~\ref{gb} after varying $k$ values in $k$-NN. In the same Fig.~\ref{gb}, the classification accuracy is high and gets affected with a small decrease after increasing the standard deviation and the value of $k$. It can be seen that RAMBP has high stability and robustness among different $k$ values. In general, RAMBP achieved the best results compared to the different state-of-the-art techniques as apparent in Tab~\ref{tab2},~\ref{tab3} and~\ref{tab4}.

\subsection{Noise-free texture classification}
\label{sec:TC}
Noise-free texture classification is challenging due to datasets properties mentioned in Tab.~\ref{tab1}. In these experiments, seven texture datasets have been used. The training and testing schemes are different from one dataset to another. For $TC10$ and $TC12$ Outex datasets, testing and training samples are well-defined by~\cite{ojala2002multiresolution}. For $TC10$ the training set has no rotation, and the testing set is rotated by \big\{$5^{\circ},10^{\circ},15^{\circ},30^{\circ},45^{\circ},60^{\circ},75^{\circ},90^{\circ}$\big\} rotation angles. Also, $TC23$ the training set has no rotation, while the testing set is rotated by \big\{$0^{\circ},5^{\circ},10^{\circ},15^{\circ},30^{\circ},45^{\circ},60^{\circ},75^{\circ},90^{\circ}$\big\} rotation angles.

$BrodatzRot$ is generated to test rotation invariance by applying a random rotation angle for each sample in $Brodatz$. For $Curet$, $Brodatz$, $BrodatzRot$, and $ALOT$ datasets, each class samples was divided equally ($50\%$ train/test) using a random selection of the samples. $100$ random couple train/test sets were generated and the classification results are averaged over the $100$ random partitionings. For $KTH2b$ dataset which has four samples of $11$ classes each, training is performed on three samples and testing on the remaining one, the results are obtained by performing the experiment four times.

The result of texture classification is depicted in Tab.~\ref{tab5}, where RAMBP provides the best results in some datasets and high performance for the others. Also, FV-CNN, SSLBP and MRELBP techniques show high and competitive performances. However, since RAMBP does not use any learning process and provides high performance for different kind of noises, RAMBP stands out the best descriptor in noisy and noise-free texture classification.

\subsection{Noisy texture retrieval}
\label{sec:IR}

Texture retrieval is based on image illustration and representation, and its basic idea is setting a query image, find the best-matched image then retrieve it~\cite{pentland1996photobook}. Texture retrieval starts by building the database, and extract its features. To retrieve a query image, the image features will be extracted and matched with the database feature vectors. Each image in the database will be ranked according to the similarity with the query image. Noisy texture retrieval is more challenging than noise-free texture retrieval, where it depends on the robustness of image features representation. So far little attention has been paid to noisy image retrieval. These experiments illustrate how the proposed method can be effective for noisy image retrieval problem. 

\setcounter{figure}{13} 
\begin{figure*}[t]
  \begin{subfigure}[b]{0.33\textwidth}
    \includegraphics[width=\textwidth]{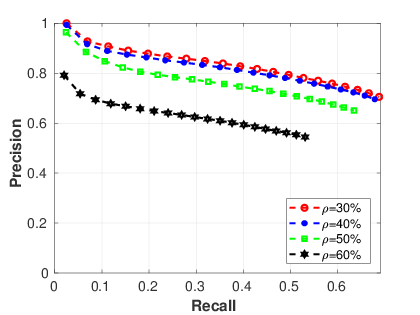}
    \caption{Salt-and-Pepper}
  \end{subfigure}
  \begin{subfigure}[b]{0.33\textwidth}
    \includegraphics[width=\textwidth]{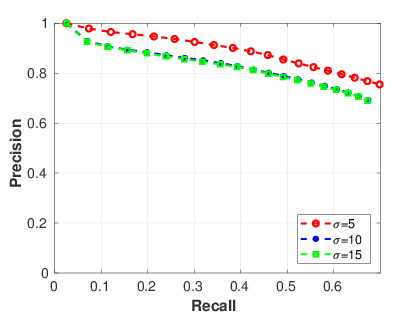} 
    \caption{Gaussian Noise}
  \end{subfigure}
  \begin{subfigure}[b]{0.33\textwidth}
    \includegraphics[width=\textwidth]{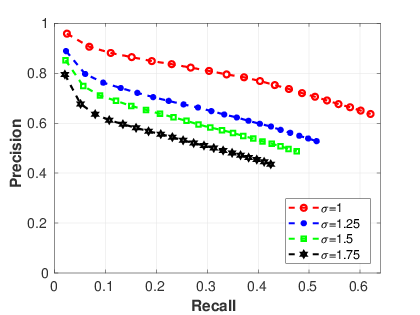}
    \caption{Gaussian Blur}
  \end{subfigure}
  \caption{RAMBP performance measure in term of average Recall and Precision for noisy texture retrieval on $Outex\_TC11$ dataset.}
 \label{fig11} 
\end{figure*}

In the literature, many types of distances have been applied. Here, ${\chi}^2$ distance is adapted to find the distances and similarities between the noisy query image and noise-free dataset images. The images are ranked in term of distances and considered as the neighbors ($k$-NN) in the feature space. This followed by returning the best similar images as the results. Fig.~\ref{fig 9} shows the flow chart of the texture retrieval process. 

\setcounter{figure}{11}
\begin{figure}[h!]
\centering
\includegraphics[width=3in]{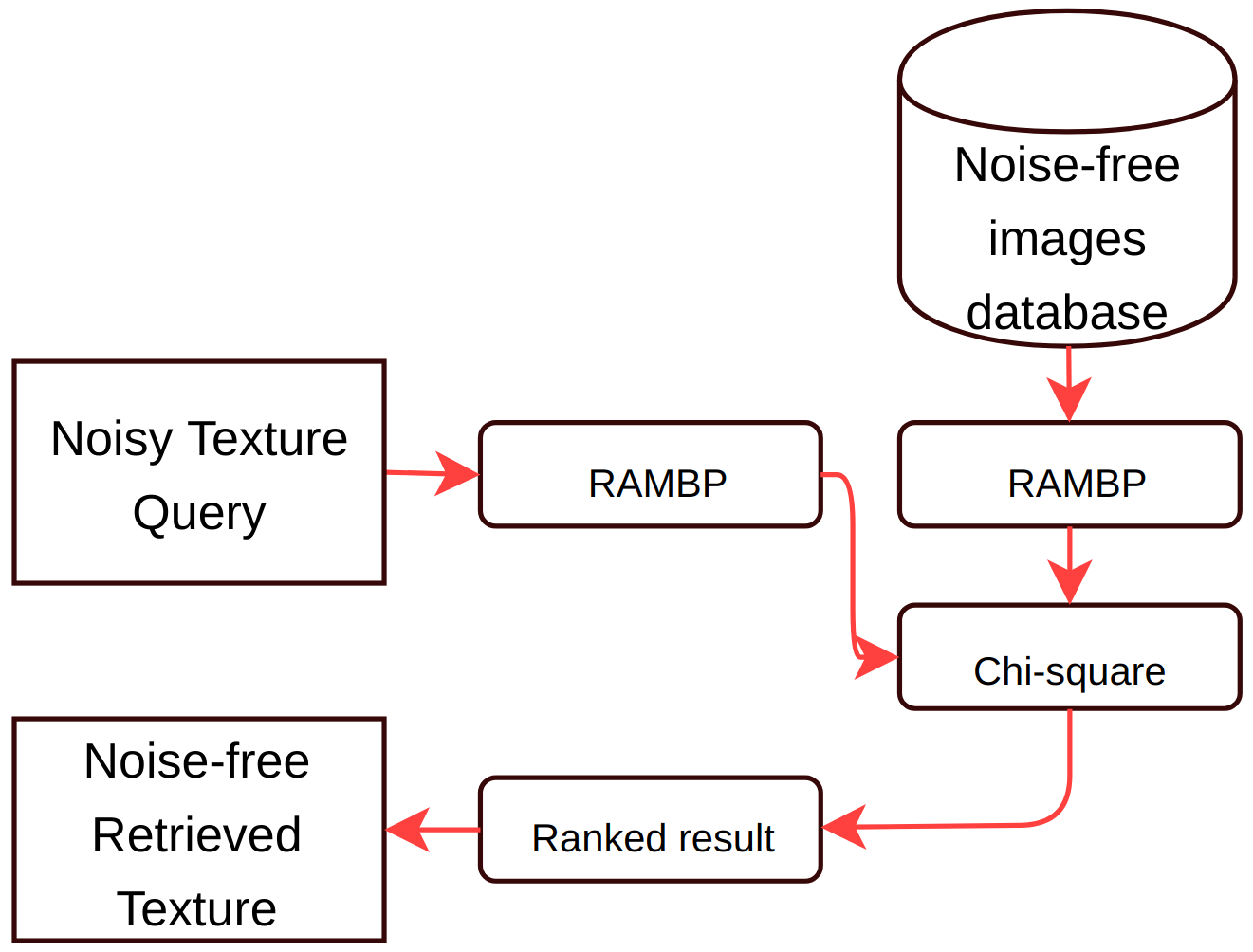}
\DeclareGraphicsExtensions.
\caption{Noisy texture retrieval scheme.}
\label{fig 9}
\end{figure}

Accuracy is the most common performance measure, but the main drawback is that accuracy hides some details that can help understanding better the classification model performance. For that, Recall and Precision provide better performance understanding by taking both false positives and false negatives into account. RAMBP noisy texture retrieval performance had been tested with regard to recall and precision by conducting the experiment on $Outex\_TC11$ dataset. Recall is the number of relevant images retrieved with respect to the number of all images in the class, while Precision is the number relevant images retrieved with respect to all retrieved images.

\begin{equation}‎\label{eq:2}
‎‎\ Recall = \frac{ Relevant\; images\; retrieved}{Total\; number\; of\; class\; images}
\end{equation}‎

\begin{equation}‎\label{eq:3}
‎‎\ Precision = \frac{ Relevant\; images\; retrieved}{Total\; number\; of\; retrieved\; images}
\end{equation}‎

In this experiment, different kinds of noise were applied to $Outex\_TC11$ database. The noisy images were chosen as the query images, while noise-free images were chosen as database images. And due to the number of images per class in $Outex\_TC11$ dataset ($40$ images per class), the number of $k$-nearest neighbor is tested up to 40 (more precise from 1 to 39 with a step of 2 to avoid tie problem). Fig.~\ref{fig 10} provides a simple example of the noisy texture retrieval ranking procedure and how to calculate the recall and precision values. 

\setcounter{figure}{12} 
\begin{figure}[h!]
\centering
\includegraphics[width=3.5in]{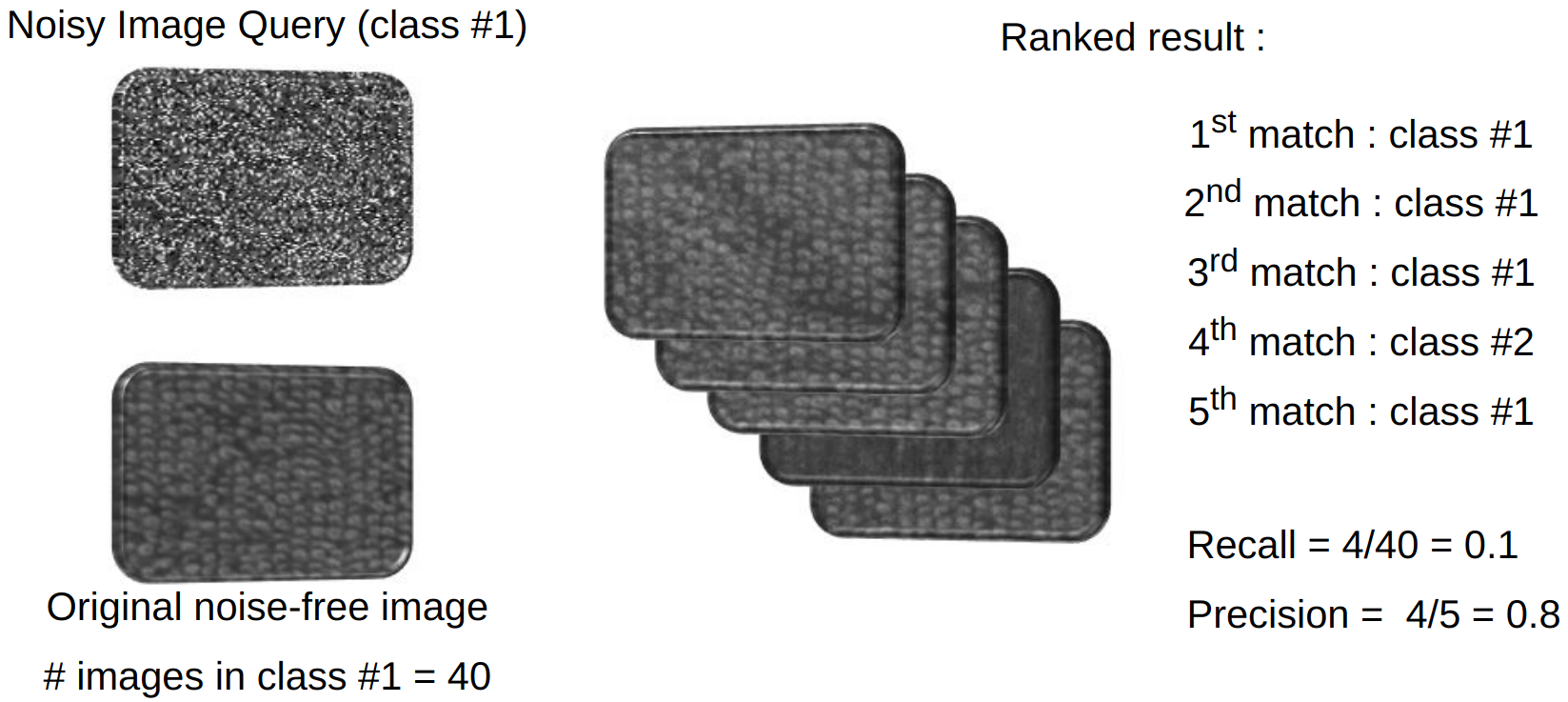}
\DeclareGraphicsExtensions.
\caption{An example of noisy texture retrieval using RAMBP with the best five results ($k=5$). In the right side, the figure depicts an example of calculating the recall and precision values according to the rank result.}
\label{fig 10}
\end{figure}

For evaluating both correctness and accuracy of RAMBP, recall and precision graph was used. And with the aim of evaluating the descriptor robustness, only textures with medium and high noises were tested. To find recall and precision values for all noisy database queries, the average of these values is taken. Fig.~\ref{fig11} provides the proposed method retrieval performance results. As it can be noticed from Fig.~\ref{fig11}, RAMBP shows high robustness and provides high recall and precision rate even under high level of noise.

\section{Conclusion}
\label{sec:Con}
Texture classification is a crucial process in many computer vision fields applications. Existing descriptors achieved good texture classification performance. However, these descriptors have some weaknesses and limitations. One of these limitations is descriptor robustness to high level of noise. Another limitation is the robustness for different kind of noises. To address these limitations Robust Adaptive Median Binary Pattern (RAMBP) descriptor is introduced in this paper. 

RAMBP descriptor takes the advantage of pixel classification and the adaptive analysis to provide strong discriminativeness and noise robustness properties. The proposed descriptor has been evaluated on noisy textures including Salt-and-Pepper, Gaussian noise, and Gaussian blur. Experimental results indicated that RAMBP outperforms other existing descriptors for handling high noisy textures classification, and performs as one of the best in noise-free texture classification. In this paper noisy texture retrieval was introduced, where it demonstrates the consistency and stability of RAMBP, and it shows the high performance and robustness of RAMBP. 

Overall, the proposed method provides the best performance in both texture classification and retrieval with a presence of very challenging different types of noise. As future work, the proposed approach will be extended to assessed on other datasets and other types of noise.
\newpage
\bibliography{mybibfile}
\end{document}